\documentclass{Interspeech}

\usepackage{arydshln}
\usepackage{multirow}
\usepackage{pifont}
\usepackage{CJKutf8} 

\interspeechcameraready

\title{WCTC-Biasing: Retraining-free Contextual Biasing ASR with Wildcard CTC-based Keyword Spotting and Inter-layer Biasing}

\author[affiliation={1,2}]{Yu}{Nakagome}
\author[affiliation={1,2}]{Michael}{Hentschel}

\affiliation{}{LINE WORKS Corporation}{Japan}
\affiliation{}{NAVER Cloud Corporation}{South Korea}
\email{y.nakagome@line-works.com}

\usepackage{comment}

\begin{document}

\maketitle

\begin{abstract}
    Despite recent advances in end-to-end speech recognition methods, the output tends to be biased to the training data’s vocabulary, resulting in inaccurate recognition of proper nouns and other unknown terms. To address this issue, we propose a method to improve recognition accuracy of such rare words in CTC-based models without additional training or text-to-speech systems. Specifically, keyword spotting is performed using acoustic features of intermediate layers during inference, and a bias is applied to the subsequent layers of the acoustic model for detected keywords. For keyword detection, we adopt a wildcard CTC that is both fast and tolerant of ambiguous matches, allowing flexible handling of words that are difficult to match strictly. Since this method does not require retraining of existing models, it can be easily applied to even large-scale models. In experiments on Japanese speech recognition, the proposed method achieved a 29\% improvement in the F1 score for unknown words.
    
\end{abstract}


\section{Introduction}
\label{sec:intro}
In recent years, the rapid progress of deep neural networks has brought about a dramatic improvement in the performance of end-to-end (E2E) automatic speech recognition (ASR) models, such as connectionist temporal classification (CTC)~\cite{Graves06_icml}, recurrent neural network transducers~\cite{Graves12_ICMLRLW}, attention-based encoder-decoders~\cite{chorowski2015attention,chan2016listen}, and decoder-only architectures~\cite{rubenstein2023audiopalm, Lakomkin24_icassp}.
These models are now widely employed in real-world applications, including conference transcription and AI dialog systems.
However, their performance heavily depends on the training data.
Consequently, it remains challenging to accurately recognize rare words such as internal jargon, personal names, and specialized technical terms when these words are rarely observed in the training corpus.

Traditional approaches have often relied on constructing or adapting decoding graphs using weighted finite state transducers (WFSTs)~\cite{zhao19d_interspeech,Huang2020ClassLA,Le2021is} to improve recognition performance for these rare words. While WFST-based decoding graph generation has proven effective, updating or rebuilding these graphs is not suitable for E2E models such as CTC, which are inherently designed to operate without decoding graphs.

In contrast, deep biasing~\cite{pundak2018deep,jain20contextual,Chang2021context,le21Contextualized,sathyendra2022contextual, huang23d_interspeech,sudo24icassp} has recently garnered attention. Deep biasing directly intervenes in the ASR training process, integrating a specified keyword list to enhance recognition of rare terms. Concretely, the approach encodes the keywords into the model’s encoder, subsequently applying bias through mechanisms such as cross-attention. However, this technique complicates model training and introduces the possibility that the set of recognizable rare words might be influenced by the keywords seen during training.

Another emerging research direction leverages speech language model (LM)~\cite{Lakomkin24_icassp, gong24b_interspeech} to enhance recognition performance.
In this approach, the LM directly accepts a context biasing list via prompts, allowing for more flexible biasing.
Nevertheless, practical constraints arise due to limitations on the number of prompt tokens. Additionally, if the acoustic model fails to correctly recognize the rare words in the first place, the biasing effect is diminished, ultimately necessitating biasing within the acoustic encoder itself~\cite{gong24b_interspeech}.

A more practical approach that requires neither model retraining nor the construction of decoding graphs is keyword boosted beam search (KBBS)~\cite{namkyu2022kwdboost}. In KBBS, beam search paths are dynamically adjusted by boosting the scores of specified keywords to encourage their appearance in the final recognition output. However, if a targeted keyword never appears in any beam hypothesis, boosting cannot be applied. This shortcoming is particularly problematic for languages such as Japanese, where there is significant variability in written forms. Moreover, CTC-based models tend to produce sharp posterior distributions~\cite{Zeyer2021WhyDC}, making it unlikely for rare words to appear among beam hypotheses. 
To address these issues, InterBiasing~\cite{nakagome24_interspeech} extends the self-conditioned CTC~\cite{nozaki2021relaxing} by injecting keywords into the intermediate layers of the acoustic encoder.
In this approach, pre-generated text-to-speech (TTS) samples are decoded to collect error patterns.
During inference, these patterns are used to correct the intermediate predictions, and the acoustic encoder is conditioned on the corrected target keywords.
While this method is effective, it requires an additional TTS module and entails a more complex decoding process.

To improve on InterBiasing~\cite{nakagome24_interspeech}, we propose \emph{WCTC-Biasing}, a novel approach that does not require any additional training or TTS modules. Specifically, during inference, we employ wildcard CTC~\cite{cai2022wctc} to compute the CTC paths for target keywords at an intermediate layer of the acoustic encoder, thereby enabling keyword detection. We then bias the acoustic encoder towards the detected keywords by conditioning the subsequent layers on these keyword candidates within the self-conditioned CTC~\cite{nozaki2021relaxing} framework.
In the acoustic encoder, wildcard CTC treats all non-keyword segments as “wildcards,” allowing efficient computation of CTC paths for only the target keywords. Moreover, because CTC utilizes both forward and backward paths, this method remains robust to pronunciation ambiguities and partial omissions. Even at lower layers of the encoder, where output hypotheses may be inaccurate, the approach can still flexibly detect keywords.
The key feature of the proposed method is that it requires neither additional training nor a TTS module, making it particularly easy to integrate into large-scale CTC-based models. 



\section{Background}
\label{sec:background}
This section describes CTC~\cite{Graves06_icml}, self-conditioned CTC~\cite{nozaki2021relaxing} and InterBiasing~\cite{nakagome24_interspeech}, which are the backbone of WCTC-Biasing.


\subsection{Connectionist Temporal Classification}
\label{sec:ctc}
E2E ASR aims to model the probability distribution of a token sequence $Y=(y_l \in \mathcal{V} \mid l=1,\dots,L)$ given a sequence of $D$-dimensional audio features $X=(\mathbf{x}_t \in \mathbb{R}^D \mid t=1,\dots,T)$, where $\mathcal{V}$ is a token vocabulary.
In the CTC framework \cite{Graves06_icml}, frame-level alignment paths between $X$ and $Y$ are introduced with a special blank token $\epsilon$.
An alignment path is denoted by $\pi=\left(\pi_t \in \mathcal{V}' \mid t=1,\dots,T\right)$, where $\mathcal{V}'= \mathcal{V} \cup \{ \epsilon \}$.
The alignment path can be transformed into the corresponding token sequence by using the collapsing function $\mathcal{B}$ that removes all repeated tokens and blank tokens.
A neural network is trained to estimate the probability distribution of $\pi_t$.
We denote the output sequence of the neural network by $Z = (\mathbf{z}_t \in (0,1)^{|\mathcal{V}'|} \mid t=1,\dots,T)$, where each element $z_{t,k}$ is interpreted as $p(\pi_t = k |X)$.
The training objective of CTC is the negative log-likelihood over all possible alignment paths with the conditional independence assumption per frame, as follows:
\begin{align}\label{eq:lossctc}
    \mathcal{L}_\mathsf{ctc}(Z, Y) = - \log \sum_{\pi\in\mathcal{B}^{-1}(Y)}\prod_{t} z_{t,\pi_t}.
\end{align}
The estimated token sequence $\hat{Y}$ is obtained as follows:
\begin{align}\label{eq:ctcdec}
    \hat{Y} = \mathcal{B}(\mathsf{argmax}(Z)).
\end{align}


\subsection{Self-conditioned CTC encoder}
\label{sec:sc-ctc}

Our model adopts an $N$-layer Conformer encoder~\cite{gulati20_interspeech} alongside the self-conditioned CTC framework~\cite{nozaki2021relaxing}. Let $X$ be the subsampled input features. The $n$-th encoder transforms its input $X^{(n-1)}$ into $X^{(n)}$:
\begin{align}
\label{eq:encoder-output}
X^{(n)} = \mathsf{Encoder}^{(n)}\bigl(X^{(n-1)}\bigr), 
\quad X^{(0)} = X.
\end{align}
After the final layer, we apply a linear projection and softmax to obtain the output sequence $Z$:
\begin{align}
\label{eq:out}
Z = \mathsf{Softmax}\bigl(\mathsf{Linear}_{D\rightarrow |\mathcal{V}'|}(X^{(N)})\bigr).
\end{align}
To improve training stability, we incorporate the idea of Intermediate CTC~\cite{lee21_icassp}. Let $Z^{(n)}$ be the softmax output at layer $n$:
\begin{align}
\label{eq:softmax}
Z^{(n)} = \mathsf{Softmax}\bigl(\mathsf{Linear}_{D\rightarrow |\mathcal{V}'|}(X^{(n)})\bigr).
\end{align}
Each $Z^{(n)}$ is used to compute the intermediate CTC loss and combined with the encoder output CTC loss as
\begin{equation}
\label{eq:loss_intermediate}
\mathcal{L}_\mathsf{ic} = (1-\lambda)\mathcal{L}_\mathsf{ctc}(Z,Y) + \frac{\lambda}{|\mathcal{N}|}\sum_{n \in \mathcal{N}} \mathcal{L}_\mathsf{ctc}\bigl(Z^{(n)},Y\bigr),
\end{equation}
where $\lambda$ is a mixing weight, $Y$ is the ground-truth sequence, and $\mathcal{N}$ indicates the encoder layers for intermediate supervision.

Self-conditioned CTC~\cite{nozaki2021relaxing} further leverages these intermediate outputs to condition subsequent layers. Specifically, each $Z^{(n)}$ is linearly mapped back to the encoder dimension $D$:
\begin{align}
\label{eq:linear-prime}
C^{(n)} &= \mathsf{Linear}_{|\mathcal{V}'|\rightarrow D}\bigl(Z^{(n)}\bigr),
\end{align}
and then added to $X^{(n)}$ to yield $X'^{(n)}$:
\begin{align}
\label{eq:selfcond}
X'^{(n)} &=
\begin{cases}
X^{(n)} + C^{(n)} & (n \in \mathcal{N}),\\
X^{(n)} & (n \notin \mathcal{N}).
\end{cases}
\end{align}
By injecting the intermediate predictions back into the encoder, the model refines its latent representations in higher layers, which improves CTC-based speech recognition performance~\cite{Higuchi21_asru}.


\subsection{InterBiasing: TTS-based keyword error collection and Inter-layer biasing}
To improve rare word recognition performance, InterBiasing~\cite{nakagome24_interspeech} has been proposed. InterBiasing, an extension of the self-conditioned CTC framework, injects the correct target keyword into the intermediate encoder layers instead of the recognition hypotheses.
This method requires generating TTS audio for each keyword $\kappa \in \mathcal{K}$, where $\mathcal{K}$ denotes the set of target keywords, as
$X_{\text{TTS},\kappa} = \mathsf{TTS}(\kappa)$, to collect misrecognitions of unknown keywords.
By applying Eqs.\ref{eq:encoder-output}, \ref{eq:softmax} to $X_{\text{TTS},\kappa}$, the intermediate predictions $\hat{Y}^{(n)}_{\text{TTS},\kappa}$ can be obtained.
The recognition hypotheses of keyword $\kappa$ at layer index $\mathcal{M}_{\mathsf{bias}}$ are selected as the trigger word set $W_{\text{trigger}, \kappa}$:
\begin{align}
W_{\text{trigger}, \kappa} = \hat{Y}^{(n)}_{\text{TTS},\kappa} \quad (n \in \mathcal{M}_{\mathsf{bias}}).
\end{align}
During decoding, if the current intermediate prediction $\hat{Y}^{(n)}$ partially matches $W_{\text{trigger}, \kappa}$, the corresponding word is replaced with $\kappa$ to form the text sequence $\hat{Y}^{(n)}_{\mathsf{bias}}$.
Next, Viterbi alignment is applied:
\begin{align}
A^{(n)}_{\mathsf{bias}} = \textsf{Viterbi}(\hat{Y}^{(n)}_{\mathsf{bias}}, Z^{(n)}),
\end{align}
and this alignment $A^{(n)}_{\mathsf{bias}}$ is converted into a one-hot vector ${Z}^{(n)}_{\mathsf{bias}}$.
Subsequently,  a weighted sum of the original softmax probability $Z^{(n)}$ and the bias one-hot vector is computed using a bias weight $\psi$:
\begin{align}
Z'^{(n)} = \mathsf{Softmax}\bigl((1 - \psi)Z^{(n)} + \psi{Z}^{(n)}_{\mathsf{bias}}),
\end{align}
and finally a linear layer transforms $Z'^{(n)}$ to obtain the bias features $C^{(n)}_{\mathsf{Bias}}$:
\begin{align}
C^{(n)}_{\mathsf{Bias}} = \mathsf{Linear}_{|\mathcal{V}'|\rightarrow D}\bigl(Z'^{(n)}\bigr).
\end{align}
This feature $C^{(n)}_{\mathsf{Bias}}$ is fed into Eq.\ref{eq:selfcond} to condition the subsequent layer.
If no trigger word is detected in $\hat{Y}^{(n)}$, the standard self-conditioned CTC is used instead.


\section{WCTC-Biasing: Wildcard CTC-based keyword spotting and Inter-layer biasing}
\label{sec:prop}

\begin{figure}[t]
  \centering
  \includegraphics[width=\linewidth]{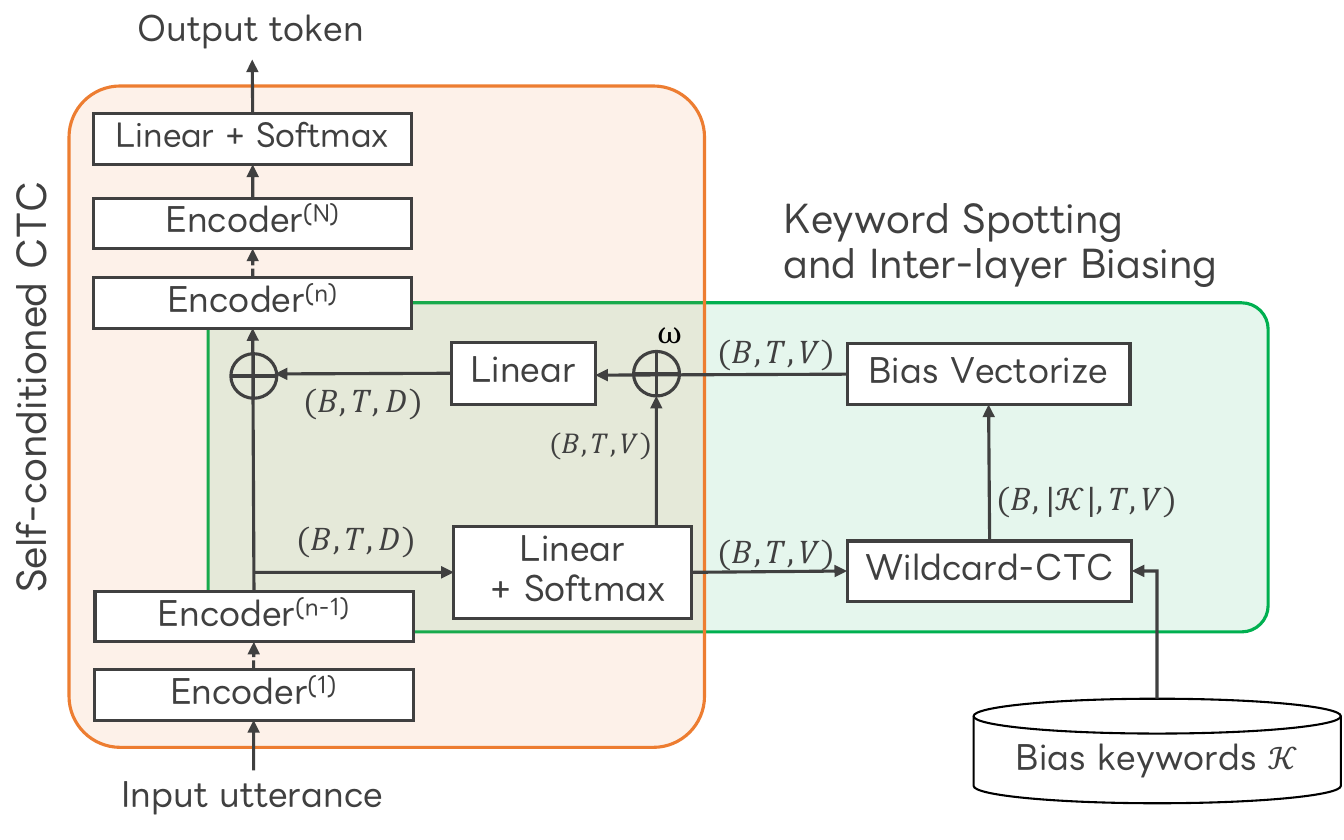}
  \caption{Overview of the proposed WCTC-Biasing. During inference, wildcard CTC is applied to an intermediate acoustic encoder layer to detect the target keyword, which is then injected into subsequent layers.}
  \label{fig:prop}
\end{figure}

Figure~\ref{fig:prop} illustrates the proposed framework. 
Keyword spotting is performed using intermediate outputs from the acoustic encoder. 
The detected keywords are converted into frame-level bias features and injected into the subsequent layers.

\subsection{Wildcard CTC-based Keyword Spotting}
\label{sec:wildcard-ctc}

Although the standard CTC framework~\cite{Graves06_icml} (Section~\ref{sec:ctc}) effectively learns a mapping between an audio feature sequence $X$ and a label sequence $Y$ by summing over all possible alignment paths, it assumes that each frame must be assigned to either a blank symbol or one of the tokens in $\mathcal{V}$. This can become problematic when certain segments (e.g., the beginning or end of an utterance) are unlabeled or deliberately omitted.
\textit{Wildcard CTC}~\cite{cai2022wctc} addresses this issue by introducing a special \emph{wildcard} token, denoted by “$\ast$.” This token can match any symbol including blank at zero cost, allowing the model to ignore unlabeled or irrelevant parts. Formally, let $\mathcal{V}' = \mathcal{V} \cup \{\epsilon\}$ be the extended CTC vocabulary, and let $\mathcal{V}'_{\ast} = \mathcal{V}' \cup \{\ast\}$ include the wildcard token. A wildcard CTC alignment path $\pi_{\ast}$ is given by
\begin{align}
\pi_\ast = (\pi_t \in \mathcal{V}'_{\ast} \mid t=1,\dots,T).
\end{align}
To encourage the model to assign the wildcard token to frames with ambiguous or low-confidence predictions, we introduce a threshold $\theta$. Specifically, we define a biased indicator for each time frame as follows:
\begin{align}
Z^{(n)}_{\mathsf{bias},\kappa, t} =
\begin{cases}
\mathsf{OneHot}_\kappa, & \text{if } \log \sum_{\pi_\ast\in\mathcal{B}^{-1}(\kappa)}p(\pi_t \mid z_t^{(n)}) > \theta, \\
0, & \text{otherwise},
\end{cases}
\end{align}
where $\mathsf{OneHot}_\kappa$ is a one-hot vector of dimension $\mathcal{V}'$ that has a value of 1 at the index corresponding to $\kappa$ and 0 in all other indices.

In summary, by allowing the wildcard token “$\ast$” to skip irrelevant frames, wildcard CTC becomes well-suited for robust keyword spotting. This is because the model can focus on whether the relevant keyword tokens appear in the encoder output sequence, without forcing all frames to be strictly labeled.

\subsection{Biasing Intermediate Predictions}
To bias the intermediate predictions toward recognizing target keywords, we interpolate the original output $Z^{(n)}$ with the aggregated biased features $Z^{(n)}_{\mathsf{bias}}$ as follows:
\begin{align}
Z^{(n)}_{\mathsf{bias}} &= \bigvee_{\kappa \in \mathcal{K}}{Z^{(n)}_{\mathsf{bias},\kappa}}, \\
Z'^{(n)} &= \mathsf{Softmax}\bigl((1-{\omega})Z^{(n)} + {\omega}Z^{(n)}_{\mathsf{bias}}),
\end{align}
where $\omega \in [0,1]$ is a weighting factor that controls the contribution of the biasing feature.
Here, the Softmax function is applied to normalize the combined scores into a valid probability distribution over the vocabulary.
Note that the biasing mechanism is applied only to the intermediate layers $n$ in the set $\mathcal{S} \subseteq \{1, \dots, N\}$, which specifies the layers selected for biasing.
For layers $n \notin \mathcal{S}$, the original predictions are used without modification.
Next, the biased intermediate predictions $Z'^{(n)}$ are converted into the biasing features:
\begin{align}
C^{(n)}_\mathsf{Bias} &= \mathsf{Linear}_{|\mathcal{V}'|\rightarrow D}\bigl(Z'^{(n)}\bigr).
\end{align}
$C^{(n)}_{\mathsf{Bias}}$ is used in Eq.\ref{eq:selfcond} to condition the subsequent encoder layer.

\section{Experiments}
\label{sec:experiment}

To evaluate the proposed method, we conducted Japanese ASR experiments using the NeMo toolkit~\footnote{https://github.com/NVIDIA/NeMo}~\cite{kuchaiev2019nemo}.
The models were evaluated based on character error rates (CERs) and F1 scores.
Following previous studies~\cite{namkyu2022kwdboost}, we used the F1 score as evaluation metric for keyword recognition.

\begin{table}[h]
\centering
\caption{Summary of keyword set sizes and average character length for OOV and IV keywords in each testset.}
\label{tab:datasets_summary}
\begin{tabular}{@{}l c c c@{}}
\toprule
& \multirow{2}{*}{Dataset} & $|\mathcal{K}|$ & Avg. char length \\
 & & OOV / IV & OOV / IV \\
\midrule
In-domain & CSJ eval3 & 23 / 8 & 4.3 / 3.1\\ \hdashline
\multirow{3}{1cm}{Out-of-domain} & Common Voice & 59 / 54 & 5.6 / 4.3 \\
 & JSUT basic 5k & 212 / 103 & 3.7 / 2.8 \\
 & TEDxJP-10K & 99 / 78 & 4.0 / 3.6 \\
\bottomrule
\end{tabular}
\end{table}

\subsection{Data}
Our model was trained on the CSJ corpus~\cite{Maekawa2003}, which consists of Japanese public speeches on academic topics.
The vocabulary $\mathcal{V}$ comprised 3,260 character units, and 80-dimensional Mel-spectrograms were used as input features.
SpecAugment \cite{Park19_interspeech} and speed perturbation \cite{Ko15_interspeech} were also applied with the ESPNet recipe~\cite{watanabe18_interspeech}.

For evaluation, we tested on one in-domain testset (CSJ eval3~\cite{Maekawa2003}) and three out-of-domain testsets (JSUT-basic 5000~\cite{Sonobe2017JSUTCF}, Common Voice v8.0~\cite{commonvoice2020}, and TEDxJP-10K~\cite{ando2021ted10k}).
The out-of-domain sets reflect typical real-world conditions where the acoustic and lexical properties of unseen user data differ from those in the training data.

Next, we describe how the bias keywords were selected. First, we decoded each evaluation set using the CSJ-trained model. By comparing the resulting hypotheses with the corresponding reference labels, we identified misrecognized words. These words were segmented using morphological analysis with MeCab \cite{kudo-etal-2004-applying}, and we retained only proper nouns and personal names consisting of two or more characters based on morphological labels. Obvious segmentation errors were then removed manually.
Finally, the remaining keywords were categorized into in-vocabulary (IV) and out-of-vocabulary (OOV) classes based on whether or not they were included in the original CSJ corpus. 
Table \ref{tab:datasets_summary} summarizes the number of IV and OOV keywords in each evaluation set.
Note that we did not use CSJ eval1 or eval2 because we could not collect a sufficient number of OOV keywords from these sets.

\begin{table*}[t]
\centering
\caption{CERs and F1 scores for Out-of-Vocabulary (OOV) and IV (In-Vocabulary) words in CSJ eval3, Common Voice, JSUT basic 5000 and TEDxJP-10K. Reported metrics are presented in the following format: CER / F1 of OOV words / F1 of IV words.“LM + BS" indicates LM shallow fusion + beam search and “LM + KBBS" indicates LM shallow fusion + Keyword-boosted beam search. $\dagger$ denotes the proposed method and $\checkmark$ indicates that a TTS module is used.}
\label{tab:methods_comparison}
\begin{tabular}{llccccc}
\toprule
& \multicolumn{6}{c}{ CER (\%) / OOV F1 / IV F1 } \\
 Methods & Decoding & TTS & CSJ eval3 & Common Voice & JSUT basic 5000 & TEDxJP-10K  \\ 
\midrule
   & Greedy & \ding{55} & 3.4 / 6.2 / 69.8 & 19.0 / 18.6 / 81.4 & 11.7 / 9.2 / 54.6 & 16.1 / 12.0 / 85.3\\
 SelfCond~\cite{nozaki2021relaxing}  & LM+BS & \ding{55}  & 3.4 / 9.2 / 72.7 & \textbf{17.3} / 18.6 / 82.8 & \textbf{11.5} / 9.2 / 54.9 & \textbf{15.8} / 12.0 / 85.6 \\ 
   & LM + KBBS &\ding{55} & 3.4 / 29.3 / 75.6  & 17.7 / 50.9 / 85.1 & \textbf{11.5} / 33.9 / 67.5 & 15.9 / 27.7 / 86.8  \\\hdashline
& Greedy& \ding{55} & 3.5 / 9.2 / 69.6 & 19.2 / 26.7 / 81.6 & 11.9 / 10.7 / 58.8 & 16.3 / 16.4 / 84.7 \\
WCTC-Biasing$\dagger$  & LM + BS& \ding{55} &  3.4 / 9.3 / 71.7 & 17.6 / 22.7 / 82.5  & \textbf{11.5} / 12.8 / 59.9  & 15.9 / 18.5 / 86.6 \\
 & LM + KBBS& \ding{55} & 3.4 / 36.3 / 80.9 & 17.9 / \textbf{57.4} / \textbf{86.5} & \textbf{11.5} / \textbf{41.6} / 69.2 & 15.9 / \textbf{35.6} / \textbf{87.6} \\\hline
& Greedy & \ding{51} & 3.5 / 27.4 / 75.6 & 19.0 / 22.7 / 82.9 & 11.7 / 13.5 / 59.5 & 16.1 / 13.1 / 85.9\\
InterBiasing~\cite{nakagome24_interspeech}  & LM + BS & \ding{51}  & 3.5 / 36.8 / 80.9 & \textbf{17.3} / 18.6 / 84.0 & \textbf{11.5} / 12.1 / 59.8 & \textbf{15.8} / 14.2 / 86.3 \\
 & LM + KBBS& \ding{51} & \textbf{3.3} / \textbf{62.4} / \textbf{83.3} & 17.7 / 52.3 / 85.8 & \textbf{11.5} / 41.5 / \textbf{70.2} & \textbf{15.8} / 33.0 / 87.4   \\ 
\bottomrule
\end{tabular}
\end{table*}

\subsection{Model Configurations}
\noindent\textbf{SelfCond:}
We adopted the Self-conditioned CTC described in Section~\ref{sec:sc-ctc}, with 18 conformer layers ($N=18$) and a hidden dimension of 512 ($D=512$).
The convolutional kernel size was 31, and the model used 8 attention heads. 
The model was trained for 50 epochs, and the final model was obtained by averaging the 10 best checkpoints based on validation CER.
The effective batch size was 120.
The Adam optimizer~\cite{Kingma14_iclr} with $\beta_1 = 0.9$, $\beta_2=0.98$,
the Noam Annealing learning rate scheduling~\cite{Vaswani17_NIPS} with 1k warmup steps was used for training.
Self-conditioning was applied at every layer ($\mathcal{N} = \{1,2,...,17\}$).

\noindent\textbf{InterBiasing:}
For generating trigger words, we synthesized speech using an in-house TTS engine. The resulting audio was passed through the SelfCond model, where greedy decoding was performed on intermediate layer outputs. We used a bias weight of $\psi = 0.9$. InterBiasing was applied to all layers from 1 to 17 ($\mathcal{M} = \{1,2,\ldots,17\}$).

\noindent\textbf{WCTC-Biasing:}
We used a bias weight of $\omega = 0.7$. For keyword spotting, the threshold $\theta$ was set to $-40$.
WCTC-Biasing was applied to every 3 layers ($\mathcal{S} = \{3,6, 9,\ldots,15\}$).

\noindent\textbf{Beam Search decoding:}
For LM shallow fusion, a 6-gram model was trained with the text corpus from the speech training data with KenLM~\cite{heafield-2011-kenlm}.
The beam size, LM weight, and length penalty were set to 10, 0.5, and 0.2, respectively, based on tuning with the CSJ dev set.
The KBBS weight was $3.0$.

\subsection{Results}
\label{sec:results} 
Table~\ref{tab:methods_comparison} summarizes the experimental results.
First, we compare the baseline model, Self-Cond~\cite{nozaki2021relaxing}, with the proposed WCTC-Biasing. 
The results indicate that, for both greedy decoding and LM beam search decoding, the proposed method consistently improves the F1 score for keywords, regardless of whether the data was in-domain or out-of-domain. 
In particular, with LM + KBBS decoding, the improvement is even more pronounced, with a relative increase of 29\% in the F1 score for OOV keywords in the TEDxJP-10K testsets. 
This improvement can be attributed to the biasing applied to the intermediate layers of the acoustic model, which causes target keywords to appear more frequently among the top candidates during beam search. 
Importantly, despite the improvement in the F1 score, the CER remains largely unchanged, suggesting that the recognition accuracy for non-keyword content is not significantly degraded.

Next, we compare the proposed WCTC-Biasing with the conventional InterBiasing approach~\cite{nakagome24_interspeech}. 
It should be noted that InterBiasing requires the use of a TTS module. The experimental results reveal that while InterBiasing achieves higher recognition accuracy on the CSJ eval3 and JSUT basic 5000 testsets, the proposed method outperforms it on the CommonVoice and TEDxJP-10K testsets.
These findings suggest that, even with a simpler architecture that does not rely on a TTS module, the proposed WCTC-Biasing can achieve recognition performance comparable to that of the conventional InterBiasing approach.
In our parameter tuning on the validation set, we also found that using a relatively low threshold $\theta$ yielded better results.
This suggests that detecting and biasing a larger number of candidate keywords increases the likelihood of capturing partially matched or weakly pronounced keywords.

\begin{figure}
  \centering
  \includegraphics[width=\linewidth]{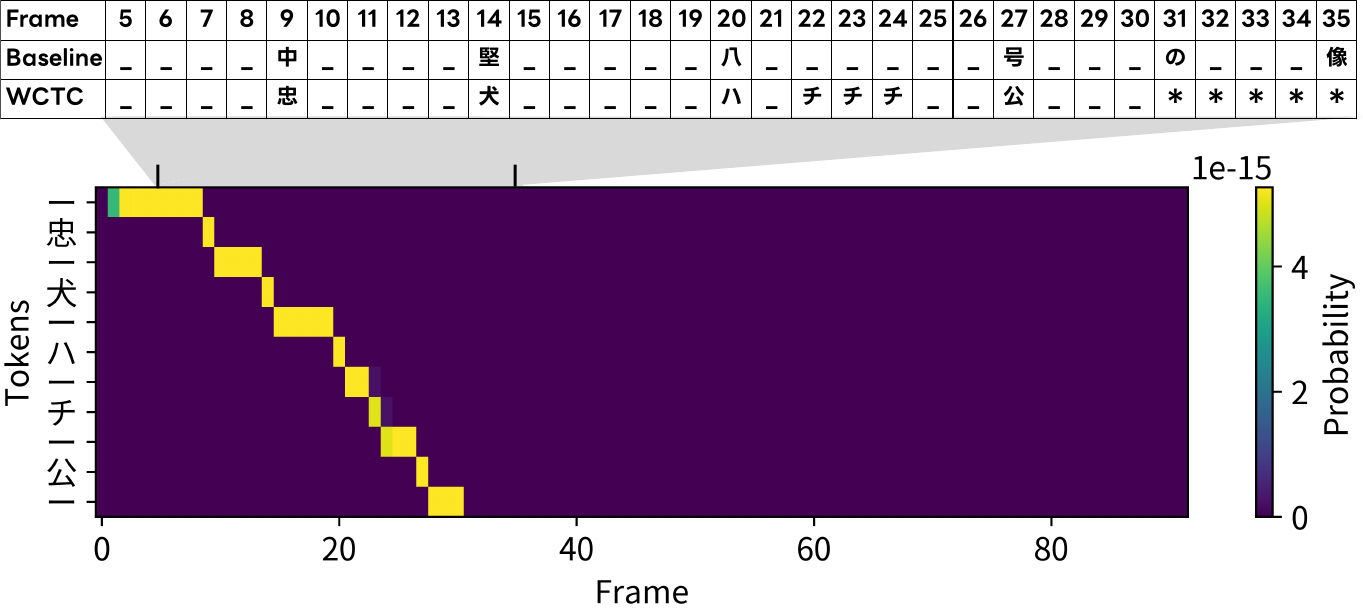}
  \caption{Wildcard CTC path for target keyword. 
  Bias keyword (OOV): “\begin{CJK}{UTF8}{min}忠犬ハチ公\end{CJK}”(Chuken Hachiko), ground truth: “\begin{CJK}{UTF8}{min}忠犬ハチ公の像は渋谷駅前に立っている\end{CJK}”, SelfCond outputs: “\begin{CJK}{UTF8}{min}中堅八号の像は渋谷駅前に建っている\end{CJK}”. “\_” is blank symbol.
  }
  \vspace{-5pt}
  \label{fig:wctc-path}
\end{figure}

Figure~\ref{fig:wctc-path} shows the alignment paths computed by the wildcard CTC for a target keyword. Although the keyword was misrecognized by the baseline Self-Cond model, it was correctly detected using wildcard CTC. The use of the wildcard allowed the model to skip non-keyword frames, and the frame range corresponding to the detected keyword was appropriately captured.

\section{Conclusions}
\label{sec:conclusions}
In this paper, we proposed a method to improve the recognition of unknown words and proper nouns in existing CTC-based models without requiring retraining. 
During inference, wildcard CTC is applied to an intermediate layer of the acoustic encoder to efficiently search CTC paths corresponding to target keywords. 
These paths are then used to bias subsequent layers. 
By treating non-keyword segments as wildcards, the proposed approach is robust to ambiguous alignments, enabling keyword detection even at lower encoder layers, where predictions are generally less reliable.
In Japanese ASR experiments, the method achieved a 29\% improvement in the F1 score for unknown words.
Since no additional training or text-to-speech modules are required, the method can be easily integrated into large-scale CTC-based ASR models, reducing operational costs while improving the recognition of target keywords.

\bibliographystyle{IEEEtran}
\bibliography{mybib}

\end{document}